\documentclass[conference]{IEEEtran}
\IEEEoverridecommandlockouts
\usepackage{url}
\usepackage{balance}
\usepackage{amsmath}
\usepackage{amssymb,bm}
\usepackage{amsfonts}
\usepackage{enumerate}
\usepackage{tabulary}
\usepackage{multirow}
\usepackage{cite}
\usepackage{lipsum}  
\usepackage{graphicx}
\usepackage{epstopdf}
\usepackage[misc]{ifsym}
\usepackage{color}
\usepackage{xcolor}
\usepackage{xxcolor}
\usepackage{pgf}
\usepackage{pgfplots}
\usepackage{tikz}
\usepackage{float}
\usepackage{lipsum}
\usepackage{tabularx} 
\usepackage[nomessages]{fp}

\definecolor{myBlue}{rgb}{0.5294, 0.8078, 0.92156}
\newlength{\maxlen}

\usepackage{array}
\usepackage{ragged2e}
\usepackage[font=footnotesize]{caption}
\usepackage{pifont}

\usepackage{colortbl}
\usepackage{svg}
\usepackage{booktabs}
\usepackage[ruled,linesnumbered]{algorithm2e}
\SetKwRepeat{Do}{do}{while}%
\definecolor{myGray}{rgb}{0.85,0.85,0.85}
\usepackage{verbatim}
\usepackage{hyperref}
\usepackage{makecell}
\usepackage{xfrac}

\usepgfplotslibrary{groupplots,fillbetween}
\usepackage{pgfplotstable}

\usepackage{ifthen}
\usepackage{forloop}
\usepackage{listings}

\definecolor{codegreen}{rgb}{0,0.6,0}
\definecolor{codepurple}{rgb}{0.58,0,0.82}
\definecolor{backcolour}{rgb}{0.95,0.95,0.92}
\lstdefinestyle{buzz}{
    backgroundcolor=\color{black!5},   
    commentstyle=\color{codegreen},
    keywordstyle=\color{blue},
    numberstyle=\tiny\color{black!30},
    stringstyle=\color{codepurple},
    basicstyle=\footnotesize\ttfamily,
    breakatwhitespace=false,         
    breaklines=true,                 
    captionpos=b,                    
    keepspaces=true,                 
    numbers=left,                    
    numbersep=5pt,                  
    showspaces=false,                
    showstringspaces=false,
    showtabs=false,                  
    tabsize=2,
}
\DeclareMathOperator*{\argmin}{argmin}

\SetAlCapSkip{0.5em}

\title{Ultra-wideband Time Difference of Arrival Indoor Localization: From Sensor Placement to System Evaluation
\thanks{* These authors contributed equally to this work. Corresponding author: wenda.zhao@robotics.utias.utoronto.ca}
} 

\author{
\IEEEauthorblockN{Wenda Zhao*}
\IEEEauthorblockA{\textit{University of Toronto}\\
Toronto, Canada}
\and
\IEEEauthorblockN{Abhishek Goudar*}
\IEEEauthorblockA{\textit{University of Toronto}\\
Toronto, Canada}
\and
\IEEEauthorblockN{Mingliang Tang*}
\IEEEauthorblockA{\textit{University of California Berkeley}\\
Berkeley, USA}
\and
\IEEEauthorblockN{Angela P. Schoellig} 
\IEEEauthorblockA{\textit{TU Munich, MIRMI}\\
Munich, Germany}
}

\begin{document}
\maketitle

\begin{abstract}

Ultra-wideband (UWB) time difference of arrival (TDOA)-based localization has emerged as a scalable positioning solution for mobile robots, consumer electronics, and wearable devices, featuring good accuracy and reliability.
While UWB TDOA-based localization systems rely on the deployment of UWB radio sensors as positioning landmarks, existing works often assume these placements are predetermined or study the sensor placement problem alone without evaluating it in practical scenarios. In this article, we bridge this gap by approaching the UWB TDOA localization from a system-level perspective, integrating sensor placement as a key component and conducting practical evaluation in real-world scenarios.
Through extensive real-world experiments, we demonstrate the accuracy and robustness of our localization system, comparing its performance to the theoretical lower bounds. Using a challenging multi-room environment as a case study, we illustrate the full system construction process, from sensor placement optimization to real-world deployment. Our evaluation, comprising a cumulative total of $39$ minutes of real-world experiments involving up to five agents and covering $2608$ meters across four distinct scenarios, provides valuable insights and guidelines for constructing UWB TDOA localization systems.

\end{abstract}


\section{Introduction}
\label{sec:intro}
\begin{figure}
    \centering
    \includegraphics[width=.4\textwidth]{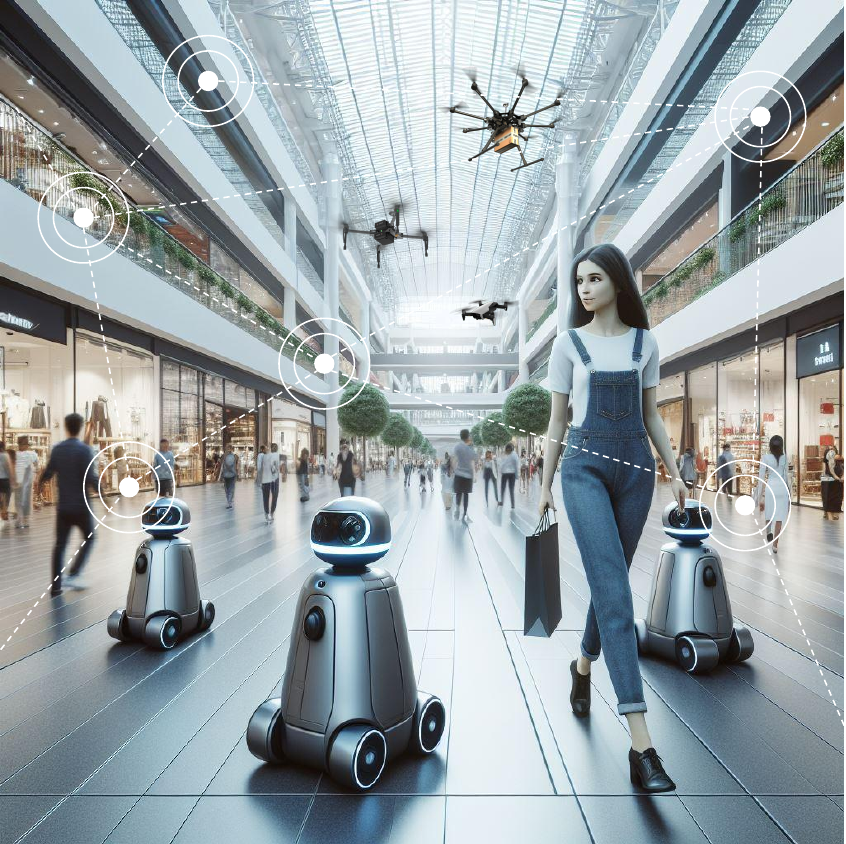}
    \caption{A conceptual diagram demonstrating the deployment of an indoor localization system in a shopping mall. The image illustrates heterogeneous agents, including ground robots and flying robots, leveraging the indoor positioning system to navigate seamlessly alongside customers while providing various services. (This conceptual image is created from an image we generated using DALL·E 3 text-to-image models developed by Open AI.)}
    \label{fig:first-image}
\end{figure}

Indoor localization technology is rapidly advancing, offering a promising future where accurate positioning enables a broad spectrum of applications, including robotics, virtual/augmented reality (VR/AR), and seamless navigation with location-based services. Precise localization and ubiquitous communication have become essential expectations for the sixth generation (6G) wireless systems~\cite{trevlakis2023localization,nabati2025opportunities}. A conceptual diagram demonstrating the indoor positioning service in a shopping mall is shown in Figure~\ref{fig:first-image}. For indoor robotics applications, visual-inertial odometry (VIO) and visual simultaneous localization and mapping (SLAM) are commonly employed techniques to achieve precise 6 degrees-of-freedom localization, with cameras providing the main source of information. However, visual localization algorithms might face challenges under dynamic lighting conditions and partial or temporary occlusion of the cameras. More importantly, capturing raw images of the surrounding environment will pose challenges for security and privacy~\cite{somasundaram2023project}. 

Compact and computationally-constrained indoor robots and smart devices have led researchers to pursue localization methods leveraging low-power and lightweight sensors. Ultra-wideband (UWB) radio technology has been shown to provide potential high-accuracy time of arrival (TOA) measurements with low power consumption compared to commonly used sensors such as camera, radar, and LiDAR in the field of robotics. UWB chips have been integrated into the latest generations of consumer electronics to support low-latency spatially-aware interactions~\cite{uwbNearbyInteration,QorvoUWB}. 



\begin{figure}[t!]
    \centering
    \includegraphics[width=.5\textwidth]{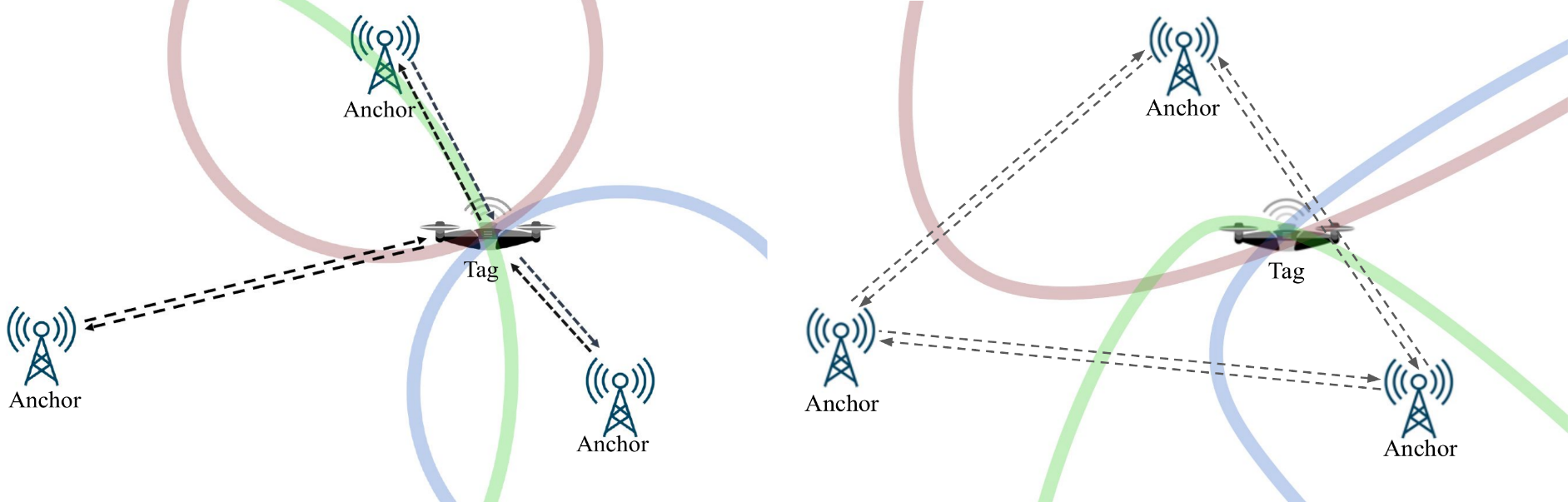}
    \caption{Conceptual diagrams for UWB TWR (left) and TDOA (right) localization system. In TWR, the UWB tag actively communicates with UWB anchors for localization. In TDOA, the UWB tag listens to the communications between anchors passively for positioning. }
    \label{fig:twr-tdoa}
\end{figure}

Similar to the Global Positioning System (GPS)~\cite{enge1994global}, UWB-based positioning systems rely on pre-installed anchors with known locations that serve as environmental landmarks. In robotics, UWB localization typically employs either two-way ranging (TWR) or time difference of arrival (TDOA) (Figure~\ref{fig:twr-tdoa}). TWR estimates distances through two-way communication between a tag and an anchor, whereas TDOA computes the arrival-time differences of signals from anchor pairs. Unlike TWR, TDOA does not require active two-way communication, enabling scalable localization of a large number of devices. This scalability is critical for applications involving many agents, such as large-scale indoor navigation and coordinated multi-robot systems. Therefore, we focus on UWB TDOA-based localization in this work.

%
Significant effort has been devoted to mitigating non-line-of-sight (NLOS) and multipath effects to improve UWB-based localization. However, most existing approaches assume fixed anchor placement and overlook its critical influence on localization performance. Prior sensor placement studies~\cite{meng2016optimal} often rely on restrictive assumptions that limit their applicability in cluttered real-world environments. In our previous work~\cite{zhao2022finding}, we proposed a more realistic sensor placement method that accounts for obstacles and multi-room scenarios, but validation was limited to simulation and controlled laboratory settings.

This article addresses this gap by presenting a system-level UWB TDOA localization framework that explicitly integrates sensor placement and evaluates its impact through real-world experiments. Using pedestrian tracking as an example, we fuse UWB TDOA and IMU measurements to estimate six-degree-of-freedom poses via an error-state Kalman filter (ESKF), and compare experimental performance against theoretical lower bounds. A challenging multi-room case study demonstrates the complete pipeline from anchor placement optimization to deployment. We additionally release our experimental dataset at \url{http://tiny.cc/uwb_tdoa_sys_dataset} to benefit the research community. To the best of our knowledge, this comprehensive evaluation, presenting a UWB TDOA-based localization system from a system-level perspective, has not been shown in the literature. 
A video summarizing our experiment process is available at \url{http://tiny.cc/uwb_tdoa_sys}. Our main contributions can be summarized as follows:
\begin{itemize}
    \item We present a system-level approach to UWB TDOA localization, incorporating sensor placement as a fundamental component. 
    
    \item We demonstrate the accuracy and robustness of the localization system through extensive experiments and compare its performance to theoretical lower bounds.
    
    \item We present the entire system deployment pipeline, from sensor placement optimization to system deployment and evaluation, through a challenging multi-room case study. 

    \item We release our experimental dataset at \url{http://tiny.cc/uwb_tdoa_sys_dataset}. 

\end{itemize}

\section{Related Work}
\label{sec:related-work}
As UWB measurements, like any other radio frequency (RF) signals, are often affected by obstacle-induced NLOS and multi-path radio propagation, multiple approaches have been proposed to improve localization accuracy under measurement corruption. M-estimators~\cite{zhang1997parameter} are commonly used as versatile tools to mitigate the influence of measurement outliers by employing robust cost functions to downweight the impact of large measurement residuals. Researchers have also explored novel models to represent UWB measurement residuals including both parametric~\cite{cano2022clock} and non-parametric models, such as neural networks~\cite{zhao2021learning}, Gaussian processes~\cite{nguyen2024third,yuan2025large}, Gaussian mixture models~\cite{zhao2023uncertainty}, and  Gaussian variational inference~\cite{stirling2026gaussian}, aiming to enhance localization performance. Furthermore, continuous-time estimation techniques~\cite{li2023continuous,goudar2023continuous} have been investigated for their applicability in asynchronous UWB-aided localization systems. 

The anchor-tag geometry, as a specific instance of the broader sensor placement problem, plays a central role in determining the performance of TDOA-based localization. In the GPS community, this effect is classically characterized using geometric dilution of precision (GDOP)~\cite{sharp2009gdop}. Cram\'{e}r-Rao lower bound (CRLB)~\cite{rao1992information} is a more general performance metric for sensor placement, which considers statistical properties of sensor measurements to evaluate estimation performance. In our previous work~\cite{zhao2022finding}, we extended the optimal sensor placement analysis to balance the effects of anchor-tag geometry and NLOS measurement biases in cluttered indoor environments. However, our experimental validation process was limited to static experiments conducted in a laboratory setup, using multilateration as the position estimator.

In this article, we present a system-level UWB TDOA localization framework that explicitly incorporates sensor placement and evaluates its impact through real-world experiments. To achieve 6 degree-of-freedom pose estimation with a low-power, lightweight design, we fuse UWB TDOA measurements with a low-cost IMU using an error-state Kalman filter (ESKF)~\cite{Roumeliotis1999}. Through a comprehensive case study in a challenging multi-room environment, we demonstrate the full pipeline from sensor placement optimization to real-world deployment, and evaluate localization performance against corresponding theoretical analysis to provide practical design guidelines.

\begin{figure*}[t!]
    \centering
    \begin{tikzpicture}
    \node[inner sep=0pt, opacity=1.0] (anchor) at (0.0,0.0)
    {\includegraphics[width=.7\textwidth]{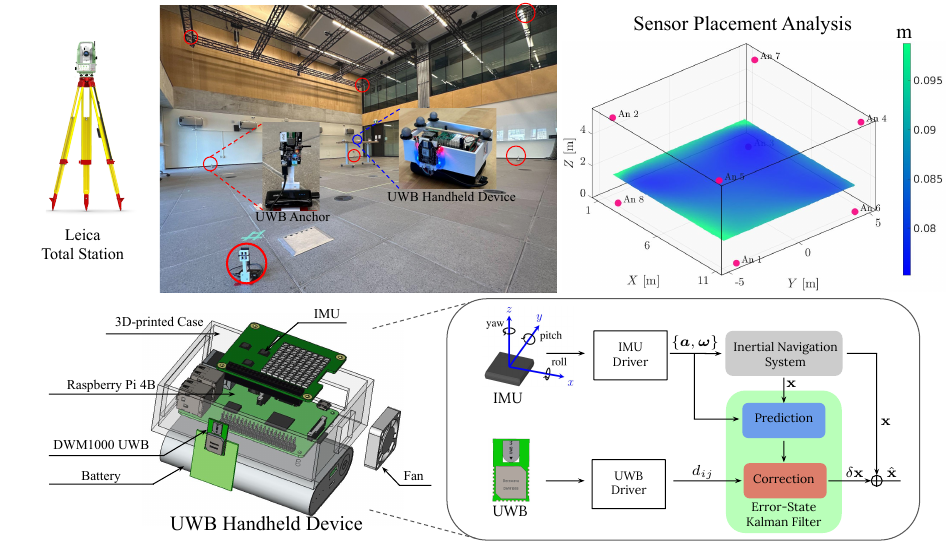}};
    \node[text width=3cm] at (-3.8,  0.0) {\small (a)};
    \node[text width=3cm] at (-2.6,  0.0) {\small (b)};
    \node[text width=3cm] at ( 2.7,  0.0) {\small (c)};
    \node[text width=3cm] at (-3.8, -3.3) {\small (d)};
    \end{tikzpicture}
    \caption{The system diagram provides a comprehensive overview of each component in our UWB TDOA localization system. The experimental setup of the multi-agent pedestrian localization  is shown in (b). UWB anchors, enclosed by red circles, are installed in the space with positions surveyed by a Leica total station (a). The heatmap in (c) illustrates the localization performance of the anchor constellation shown in (b), calculated at a height of $1.5$ meters using sensor placement analysis. Lower root-mean-squared error (RMSE) is indicated as darker color. The hardware components of our UWB handheld device along with the onboard ESKF localization algorithm are shown in (d). 
    }
    \label{fig:system-setup}
\end{figure*}
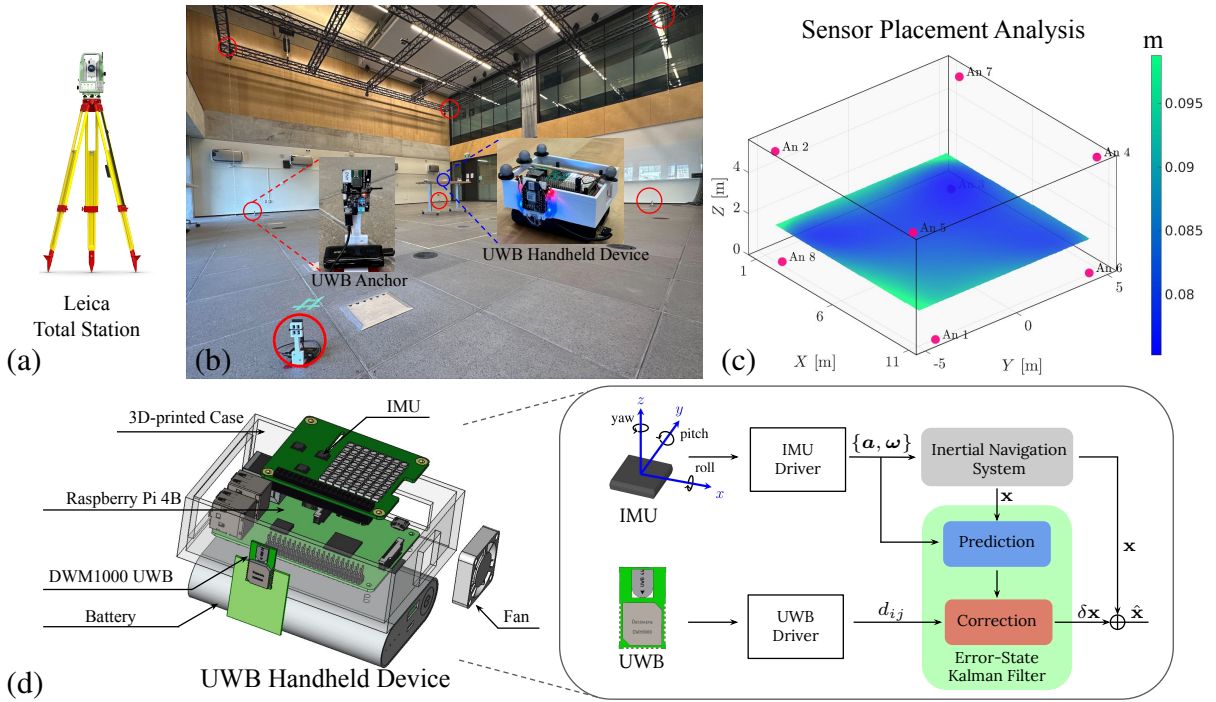

\section{UWB TDOA Localization System}
\label{sec:uwb-system}
In the following section, we provide an overview of our UWB TDOA localization system. We first introduce the sensor placement optimization and the UWB TDOA-IMU localization algorithm. Then, we present the hardware and software architecture of our UWB TDOA localization system as well as the details of system deployment. The overall system architecture is shown in Figure~\ref{fig:system-setup}. 
\subsection{Preliminaries and Notation}
A general UWB TDOA-based localization system consists of a set of $m$ UWB anchors, divided into anchor pairs $\Gamma = \{(1,2), (2,3), \cdots, (m-1, m)\}$ that are pre-installed in an indoor space $\mathcal{P} \in \mathbb{R}^{3}$. To facilitate our analysis, we define a vector $\bm{\alpha}=\left[\bm{a}_1^T, \bm{a}_2^T, \cdots, \bm{a}_{m}^T\right]^T \in \mathbb{R}^{3\cdot m}$ to denote the anchor placement. Five handheld devices are designed for localization purposes, each equipped with one IMU and one UWB tag. We refer to the absolute coordinate frame created by the UWB anchors as the inertial frame~$\mathcal{F}_{\mathcal{I}}$ and denote the handheld device body frame as~$\mathcal{F}_{\mathcal{B}}$. 

\subsection{Sensor Placement Optimization}
\label{sec:osp}
Sensor placement optimization is often conducted through the lens of statistical parameter estimation, which optimizes the performance bounds for estimating the parameters of interest with a given model. Considering the UWB TDOA localization system in particular, we analyze the lower bound on the position estimate based on the UWB anchor placement and the TDOA measurement model. This analysis provides insights into the fundamental limitations of the system's localization accuracy. In our previous work~\cite{zhao2022finding}, we leverage the mean-squared error (MSE) metric to evaluate the UWB TDOA localization performance for a region of interest in cluttered environments. The estimated tag position $\hat{\mathbf{p}}$ is characterized by its bias $\textrm{Bias}(\hat{\mathbf{p}})$ and covariance matrix $\textrm{Cov}(\hat{\mathbf{p}})$. The MSE of an estimate $\hat{\mathbf{p}}$ of the true value $\mathbf{p}$ can be decomposed as 
\begin{equation}
    \textrm{MSE}(\hat{\mathbf{p}}) = \mathbb{E}\{\|\hat{\mathbf{p}} - \mathbf{p}\|^2\} = \textrm{Tr}\left(\textrm{Cov}(\hat{\mathbf{p}})\right) + \|\textrm{Bias}(\hat{\mathbf{p}})\|^2,
\end{equation}
where $\textrm{Tr}(\cdot)$ is the trace operator and  $\|\cdot\|$ is the $\ell_2$ norm. Based on the uniform Cram\'{e}r-Rao bound~\cite{eldar2008rethinking} and linear approximation, we can compute the lower bound of the $\textrm{MSE}(\hat{\mathbf{p}})$, denoted as $M(\mathbf{p}, \bm{\alpha})$, with respect to a given anchor placement $\bm{\alpha}$. For a region or a trajectory of interest, we evaluate the MSE metric at $N$ sample points $\mathbf{p}_i \in \Phi, i=1\cdots N$, and compute the average root-mean-squared error (RMSE) 
\begin{equation}
\label{eq:mse_metric}
    \mathcal{M}(\bm{\alpha}) = \frac{1}{N}\sum_{i=1}^{N} \sqrt{M(\mathbf{p}_i, \bm{\alpha})}
\end{equation}
as the performance metric of the sensor placement $\bm{\alpha}$.

\begin{figure*}[t!]
    \centering
    \includegraphics[width=.7\textwidth]{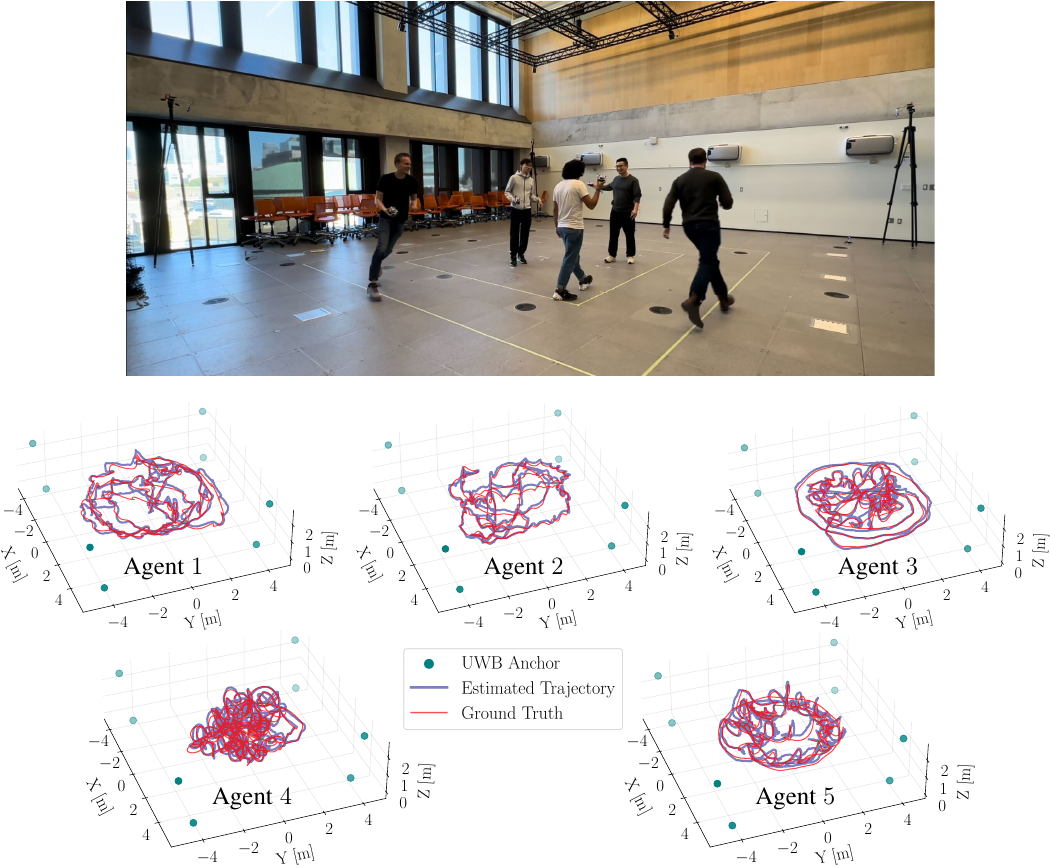}
    \caption{The multi-agent pedestrian localization experiment with five agents in constellation $1$ is shown in the photo above. The estimated together with the ground truth trajectories of each agent during the experiments are shown in the bottom plots. Readers are encouraged to view our supplementary video (\url{http://tiny.cc/uwb_tdoa_sys}) to gain a better insight into our experimental process and evaluate the robustness of the localization performance.     
    }
    \label{fig:multi-agent-local}
\end{figure*}
Optimal anchor placement is essential for ensuring reliable UWB TDOA localization performance, particularly in cluttered and geometrically challenging environments. Therefore, it is important to optimize the anchor positions for regions or trajectories of interest during system design. We define a set $\mathcal{A}$ containing all possible anchor configurations with $\bm{\alpha}\in \mathcal{A}$ and use a block coordinate-wise minimization (BCM) algorithm to find the optimal placement of UWB anchors $\bm{\alpha}^{\star}$ that minimizes $\mathcal{M}(\bm{\alpha})$:
\begin{equation}
\label{eq:optimization_problem}
    \bm{\alpha}^{\star} = \argmin_{\bm{\alpha}\in \mathcal{A}} \mathcal{M}(\bm{\alpha}).
\end{equation}
Readers are encouraged to refer to~\cite{zhao2022finding} for a comprehensive explanation of the aforementioned optimal sensor placement algorithm.

\subsection{UWB TDOA-Inertial Navigation System}
To achieve 6 degrees-of-freedom pose estimation, we fuse UWB TDOA measurements with an IMU sensor, leading to a UWB TDOA-inertial navigation system. We follow the parameterization from~\cite{goudar2021online} and describe the system with a $16$-dimensional state vector:
\begin{equation}
\mathbf{x}(t) = (\mathbf{p}(t), \mathbf{v}(t), \mathbf{q}_{\mathcal{IB}}(t), \mathbf{b}_a(t), \mathbf{b}_\omega(t)), \label{eqn:state}
\end{equation}
where $\{ \mathbf{p}(t), \mathbf{v}(t), \mathbf{q}_{\mathcal{IB}}(t) \}$ denote the position, linear velocity, and orientation of the IMU body frame with respect to the inertial frame. A unit quaternion parameterization is used for representing orientations. Accelerometer and gyroscope biases are denoted by $\mathbf{b}_a(t)$ and $\mathbf{b}_\omega (t)$.  The spatial offset, also called lever-arm, between the IMU and UWB tag is denoted by $\bm{l}_{ub}$.

The measured angular rate by a gyroscope $\bm{\omega}_m = (\omega_x, \omega_y, \omega_z)$ is related to the true angular rate $\bm{\omega}_t$ as: $\bm{\omega}_m = \bm{\omega}_t + \mathbf{b}_\omega + \mathbf{n}_\omega$, where $\mathbf{b}_\omega$ is the time-varying bias and $\mathbf{n}_{b\omega}$ is a zero-mean additive white Gaussian noise (AWGN) process with covariance $\mathbf{Q}_\omega$, i.e. $\mathbf{n}_\omega \sim \mathcal{N}(\mathbf{0}, \mathbf{Q}_\omega)$. The bias is modelled as driven by another AWGN process $\mathbf{n}_{b\omega} \sim \mathcal{N}(\mathbf{0}, \mathbf{Q}_{b\omega})$: $\dot{\mathbf{b}}_\omega = \mathbf{n}_{b\omega}$. A similar model is used for the accelerometer.

To facilitate the localization of multiple heterogeneous agents, the motion model used in this work is a 3D kinematic motion model with IMU measurements as inputs~\cite{goudar2021online}. The relevant equations and a detailed description of the motion model can be found in~\cite{goudar2021online}. 
The UWB TDOA measurements complement the motion dynamics by providing drift-free difference-of-distance measurements between UWB anchor pairs and the UWB tag. The measurement model for an anchor pair $\{\bm{a}_i,\bm{a}_j\}$ is described as follows: 
\begin{equation}
\small
d_{ij,t} = \|\mathbf{C}_{\mathcal{IB}}(t) \bm{l}_{ub} + \mathbf{p}(t) -\bm{a}_j\| - \|\mathbf{C}_{\mathcal{IB}}(t) \bm{l}_{ub} + \mathbf{p}(t) -\bm{a}_i\| + \eta_{ij}(t), 
\label{eqn:meas_model}
\end{equation}
where $\mathbf{C}_{\mathcal{IB}}(t):=\mathbf{C}\{\mathbf{q}_{\mathcal{IB}}(t)\}$ represents the rotation matrix from the body frame to the inertial frame. The UWB measurement noise, denoted as $\eta_{ij}(t)$, is modeled to follow a zero-mean Gaussian distribution $\eta_{ij}\sim\mathcal{N}(0,\sigma_{ij}^2)$ and is assumed to be common to all TDOA measurements.

To estimate the system state, we employ an error-state Kalman filter (ESKF)~\cite{Roumeliotis1999}. In the prediction step, inertial dead reckoning propagates the state (\ref{eqn:state}) using a 3D kinematic motion model driven by IMU measurements, while the associated uncertainty is propagated accordingly. In the correction step, UWB TDOA measurements are used to estimate the error between the predicted state and the measurement-consistent state. The corresponding error state is defined as
\begin{equation}
\delta \mathbf{x} = (\delta \mathbf{p}, \delta \mathbf{v}, \delta \bm{\theta}, \delta \mathbf{b}_a, \delta \mathbf{b}_{\omega}),
\label{eqn:error_state}
\end{equation}
where $\delta \mathbf{p}$ and $\delta \mathbf{v}$ denote position and velocity errors, $\delta \bm{\theta}$ is the local orientation error, and $\delta \mathbf{b}_a$ and $\delta \mathbf{b}_{\omega}$ represent accelerometer and gyroscope bias errors, respectively. The orientation error is parameterized via a small-angle quaternion $\delta \mathbf{q} = (1, \tfrac{1}{2}\delta \bm{\theta})$, with $|\delta \bm{\theta}| \ll 1$.

The estimated error is then composed with the nominal state to correct accumulated drift, $\hat{\mathbf{x}} = \mathbf{x} \oplus \delta \mathbf{x}$, where $\oplus$ denotes quaternion multiplication for orientation and additive updates for the remaining states. Further details on the ESKF prediction and correction steps can be found in~\cite{Sola2017}.

\subsection{System Architecture and Deployment Details}
\label{sec:arch_deploy}
To evaluate localization performance, we assembled five handheld UWB devices (Figure~\ref{fig:system-setup}d) using the Bitcraze Loco Positioning System (LPS), which employs low-cost DWM1000 UWB radios for TDOA localization. Each device integrates a DWM1000 UWB tag, a low-cost LSM9DS1 IMU, and a Raspberry Pi 4 Model B (8 GB RAM) powered by a portable battery, all enclosed in a 3D-printed case. All system components are configured as plug-and-play to facilitate deployment. The handheld devices run Ubuntu 20.04 LTS with ROS Noetic, and the ESKF-based localization algorithm is implemented in C++, operating at $420$ Hz while consuming approximately $35\%$ of a single CPU core and $0.4\%$ of system memory.

During deployment, a millimeter-accurate Leica total station (Figure~\ref{fig:system-setup}a) is used to establish the inertial frame and survey anchor positions. Using the “Orientate to line” procedure, the inertial frame is defined by two surveyed points, after which all anchors are measured within this frame with millimeter accuracy. The UWB localization system then provides drift-free positioning relative to the established inertial frame.

\section{Experiments}
\label{sec:experiment}
We evaluate the proposed UWB TDOA localization system through extensive real-world experiments across multiple environments. We first demonstrate multi-agent pedestrian localization and quantify the impact of anchor placement using two anchor constellations. We then showcase the system's ability to provide reliable positioning under sensor occlusion. Next, we show how sensor placement analysis captures localization degradation caused by challenging anchor-tag geometry in a staircase environment. Finally, using a complex multi-room scenario consisting of a cafeteria and a narrow hallway as a case study, we demonstrate the full system pipeline—from sensor placement optimization to deployment—and compare experimental results against theoretical lower bounds. We release our experimental dataset together with development kit at \url{http://tiny.cc/uwb_tdoa_sys_dataset}. A summary video is available at \url{http://tiny.cc/uwb_tdoa_sys}.
%
\begin{figure}[t!]
    \centering
    \includegraphics[width=0.45\textwidth]{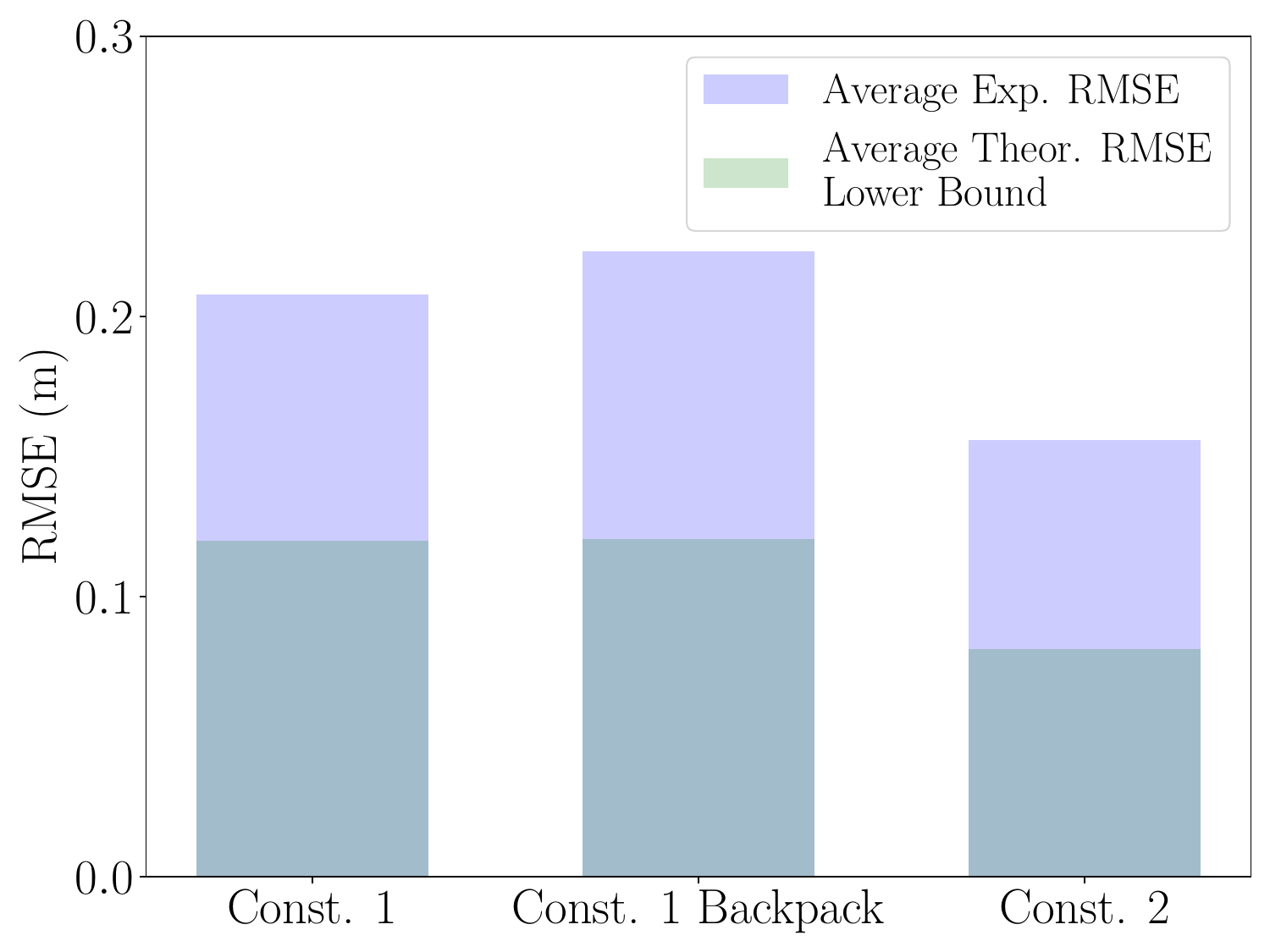}
    \caption{Comparison of the average experimental root-mean-square error (RMSE) for multi-agent pedestrian localization in Const. $1$ and $2$, along with localization under occlusion in Const. $1$ (Const. 1 Backpack), shown as blue bars, against the theoretical RMSE lower bounds represented by green bars. 
    }
    \label{fig:barplot}
\end{figure}
\subsection{Multi-agent Pedestrian Localization}
We first evaluate the accuracy and robustness of our UWB TDOA positioning system in multi-agent pedestrian localization through comprehensive real-world experiments. Eight DWM1000 UWB radios are deployed as anchors in centralized TDOA mode with a round-robin topology $\Gamma=\{(8,1),(1,2),\dots,(7,8)\}$. A detailed explanation of centralized and decentralized TDOA modes using Bitcraze's UWB anchors can be found in~\cite{zhao2024util}. The anchors are placed at the corners of an open two-story indoor flight arena to maximize coverage (Figure~\ref{fig:system-setup}b). The UWB TDOA noise standard deviation is set to $\sigma_{ij}=0.1$~m, and a chi-squared test with Mahalanobis threshold $5$ is used for outlier rejection. Although the UWB TDOA localization system can provide positioning services to an unlimited number of agents, we demonstrate the multi-agent pedestrian localization performance using five identically designed handheld devices in our experiments.

Two anchor constellations are evaluated. In the first constellation, four anchors are mounted on tripods at a height of $2.65$~m. Five participants freely walked and ran within the anchor convex hull over three trials. We demonstrate one trial of our experiments together with a comparison of the estimated and the ground truth trajectories for each agent in Figure~\ref{fig:multi-agent-local}. Across the three experimental trials, our localization system demonstrated consistent pedestrian localization performance for the five agents, achieving an average RMSE of $21$~cm with a small standard deviation of $1.56$~cm. In the second anchor constellation, we mounted the four anchors on the room's ceiling frame, increasing the z-axis separation to $5.5$~m. Three participants carried the handheld devices and conducted three trials of experiments by walking around the same room. With identical parameters applied in the ESKF algorithm, the average RMSE for the three experiments in the second constellation is $16$~cm  with a similar standard deviation of $1.63$~cm.  

\begin{figure}[b!]
    \centering
    \includegraphics[width=.43\textwidth]{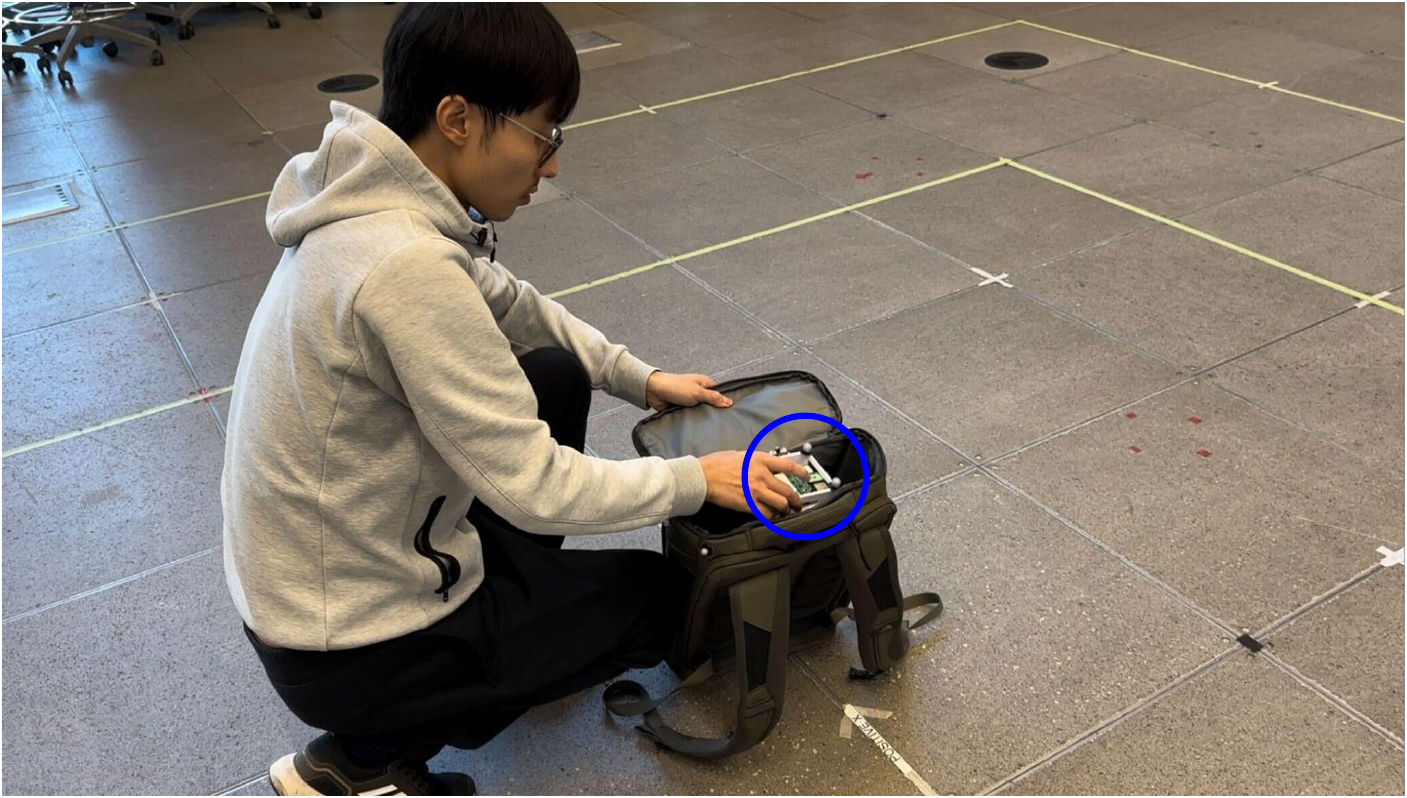}
    \caption{The UWB handheld device, highlighted by a blue circle, was put into a backpack to demonstrate the localization performance under sensor occlusion.}
    \label{fig:occluded-local}
\end{figure}
Through these comparative experiments, we observe that changing the sensor placement alone leads to a notable impact on the localization performance, which highlights the importance of UWB anchor placement. We calculate the theoretical lower bounds of localization RMSE in the aforementioned two anchor constellations using sensor placement analysis. The average theoretical RMSE lower bounds in the first and second anchor constellations are $12$~cm and $8$~cm, respectively. We summarize the average experimental RMSE values for multi-agent pedestrian localization and the theoretical RMSE lower bounds in Figure~\ref{fig:barplot} for comparison. The experimental results agree with the theoretical analysis that sensor placements with lower theoretical RMSE bounds generally yield better localization performance.  

\subsection{Localization under Sensor Occlusion}
One distinctive feature of UWB-Inertial localization is that the sensor set can be partially or completely occluded thanks to the obstacle-penetrating ability of UWB measurements. To demonstrate this feature, we put one UWB handheld device into a backpack (see Figure~\ref{fig:occluded-local}) and conducted three trials of experiments in the aforementioned first anchor constellation. Since the device was completely occluded by the backpack throughout the experiments, we created a motion capture tracking object for the backpack to record the ground truth position data approximately. The average localization RMSE together with the theoretical lower bound is summarized in Figure~\ref{fig:barplot}. 

The average RMSE from the three experiments is $22$~cm, which closely matches the performance of multi-agent pedestrian localization ($21$~cm) within the same constellation. Although the fabric backpack did not pose a severe NLOS challenge for UWB measurements, conventional localization algorithms relying on cameras or LiDARs are unsuitable for this application. This experiment demonstrates that the UWB-inertial localization system has the potential to provide reliable positioning performance even under complete sensor occlusion.

\begin{figure}[b!]
    \centering
    \begin{tikzpicture}
    \node[inner sep=0pt, opacity=1.0] (anchor) at (0.0,0.0)
    {\includegraphics[width=.475\textwidth]{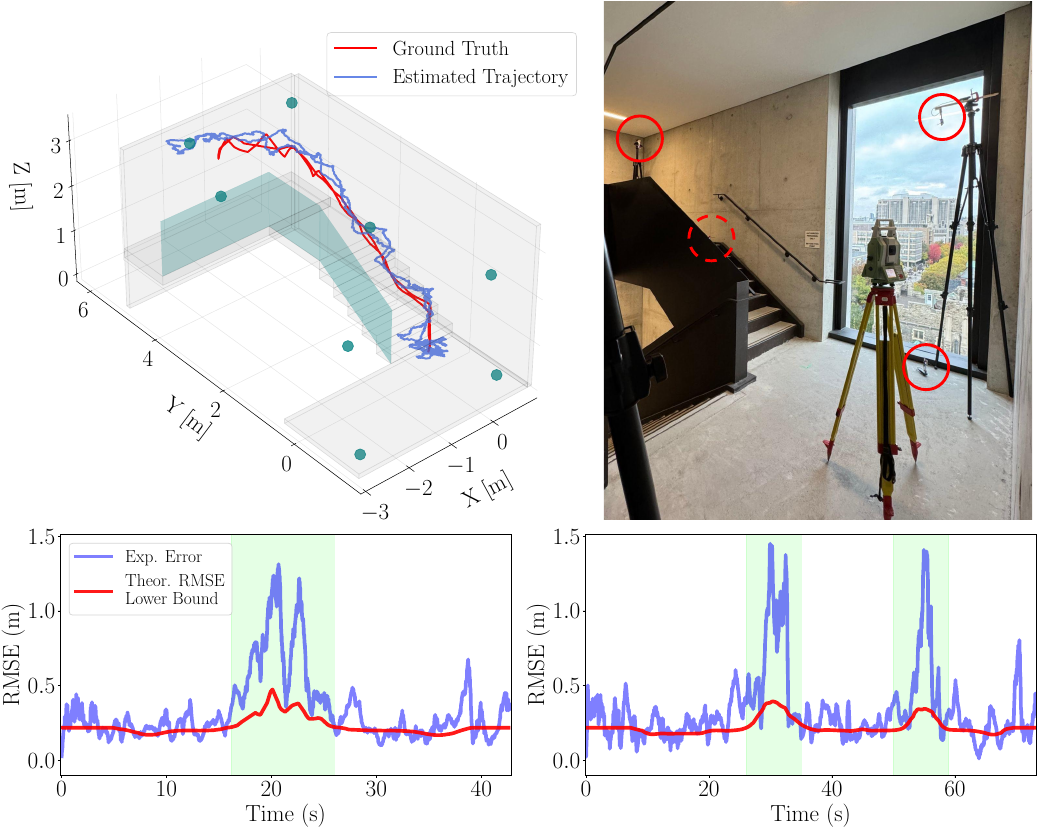}};
    \node[text width=3cm] at (-2.8, 3.4) {\normalsize (a)};
    \node[text width=3cm] at ( 1.83, 3.4) {\normalsize (b)};
    \node[text width=3cm] at (1.5, -3.5) {\normalsize (c)};
    \end{tikzpicture}
    \caption{The staircase scenario and our experimental setup are shown in (a) and (b). The experimental trajectory $\#1$ and the corresponding ground truth trajectory are shown in (a). The ESKF estimation errors and the theoretical RMSE lower bounds along the trajectories are shown in (c).}
    \label{fig:stair-setup}
\end{figure}
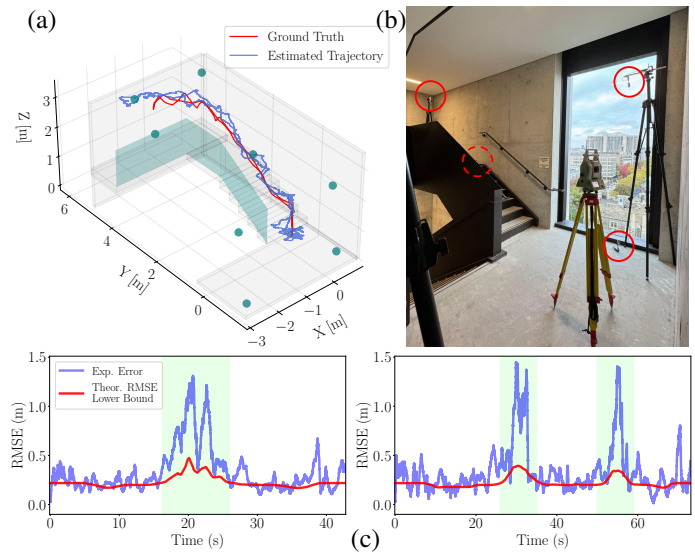
\subsection{Localization in a Staircase Environment}
\label{sec:stair_exp}
Realistic working conditions, such as building monitoring or multi-room scenarios, often pose geometric challenges and lead to degradation in UWB TDOA localization performance. Therefore, evaluating localization performance through sensor placement analysis is crucial to avoid areas with poor positioning accuracy. To showcase this ability, we conducted experiments in a geometrically challenging staircase environment. We illustrate how sensor placement analysis effectively captures the degradation of localization accuracy induced by the difficult anchor-tag geometry in staircase positioning scenarios.

\begin{figure}[t!]
    \centering
    \begin{tikzpicture}
    \node[inner sep=0pt, opacity=1.0] (anchor) at (0.0,0.0)
    {\includegraphics[width=.5\textwidth]{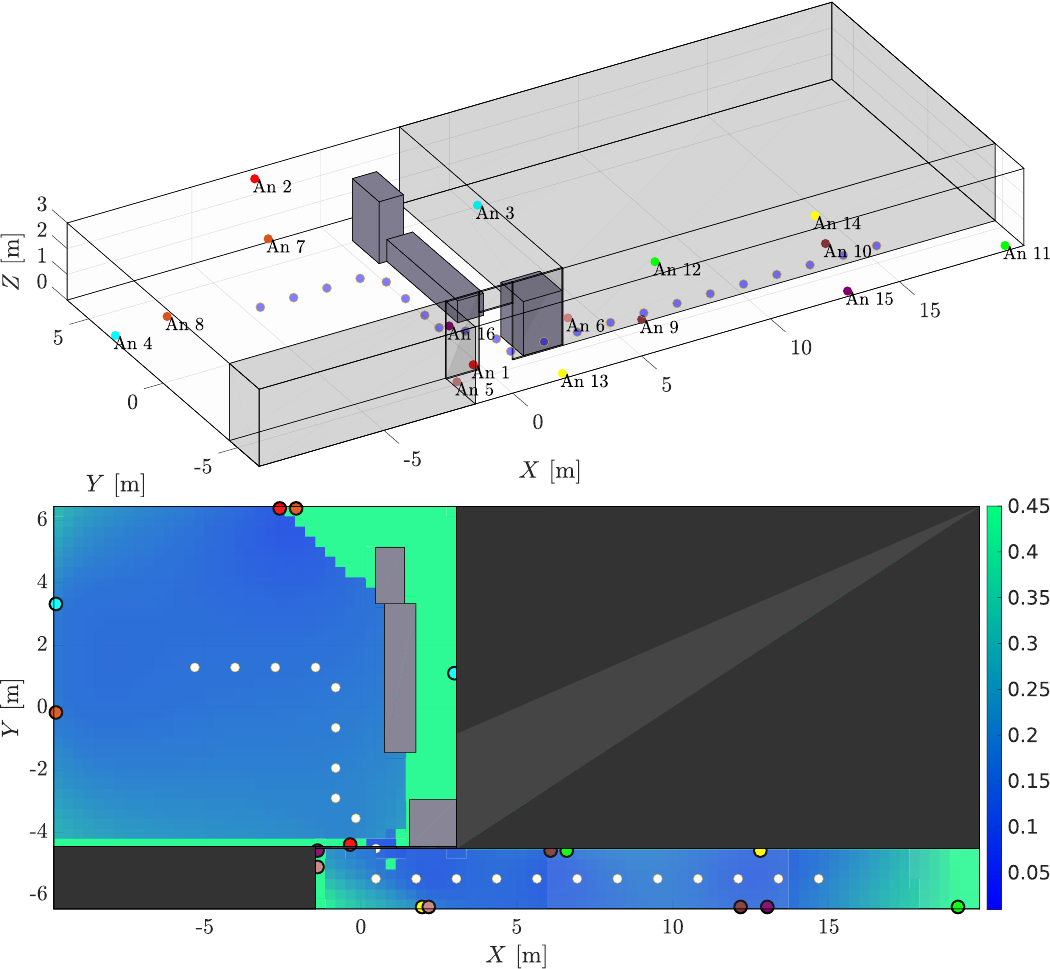}};
    \node[text width=3cm] at (-2.8, 0.7) {\normalsize (a)};
    \node[text width=3cm] at (-2.8, -4.0) {\normalsize (b)};
    \end{tikzpicture}
    \caption{The 3D layout of the multi-room environment, including a cafeteria and a narrow hallway, is shown in (a). The $22$ sampled points representing the pre-defined trajectory are shown as blue dots in (a) and white dots in (b). The $16$ UWB anchors in decentralized TDOA mode are designed through sensor placement optimization. The localization performance among the multi-room space is demonstrated in the heatmap (b), with lower RMSE indicated with darker color.} 
    \label{fig:hallway-design}
\end{figure}
The staircase scenario and our experimental setup are shown in Figures~\ref{fig:stair-setup}a and b. Eight anchors were deployed in centralized TDOA mode to cover the staircase while maintaining inter-anchor line-of-sight. In the ESKF algorithm, we increased the variance of UWB measurement noise to $\sigma_{ij}^2=0.015$ to account for degraded radio conditions. We use a Leica total station in the tracking mode, which tracks the prism on the UWB handheld device and provides the position measurement at $5$~Hz for ground truth. To prevent the total station from losing track of the prism, we conducted two low-speed experiments walking along the staircase to evaluate the localization performance. We demonstrate the ground truth trajectory of the first experiment along with the estimation results in Figure~\ref{fig:stair-setup}a. During the experiments, we moved the handheld device upstairs to the second floor and approached the boundary of the anchor constellation, which resulted in challenging anchor-tag geometry~\cite{zhao2022finding}. 

\begin{figure*}[t!]
    \centering
    \begin{tikzpicture}
    \node[inner sep=0pt, opacity=1.0] (anchor) at (0.0,0.0)
    {\includegraphics[width=.75\textwidth]{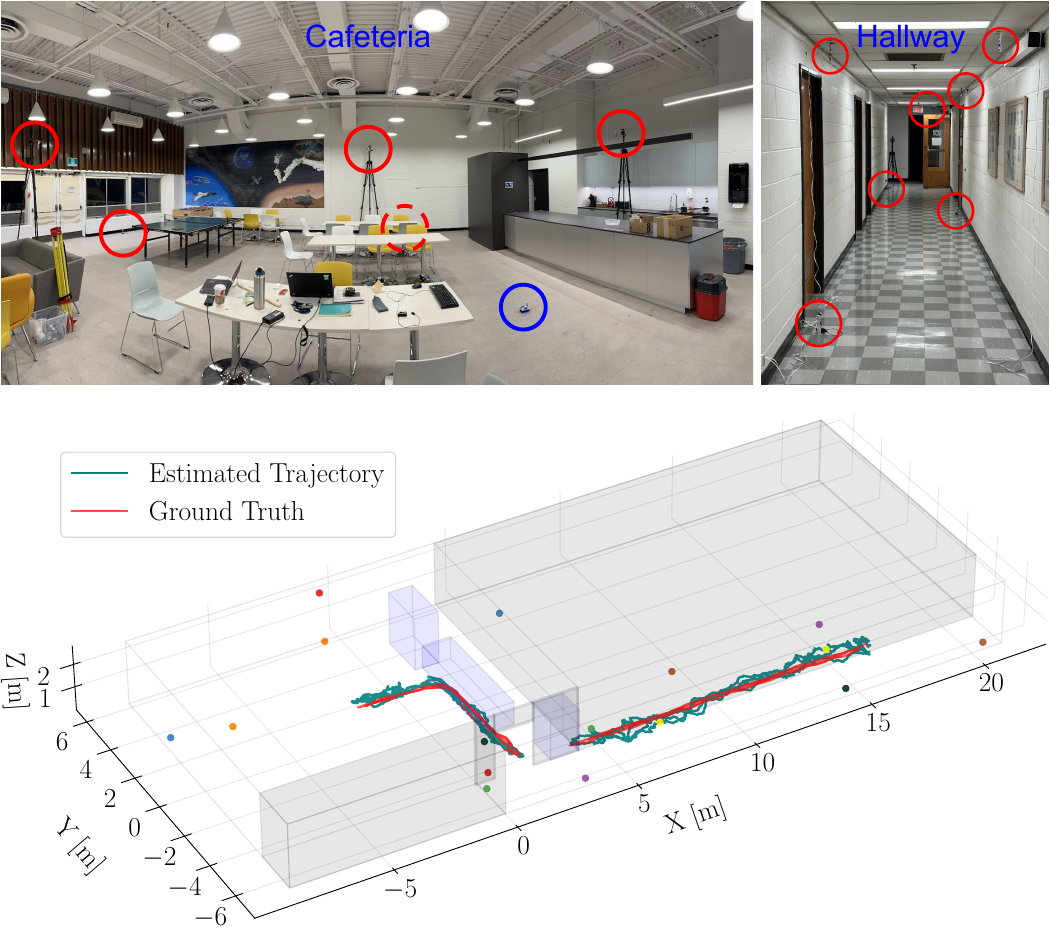}};
    \node[text width=3cm] at (-6.0, 1.0) {\large (a)};
    \node[text width=3cm] at (-6.0, -5.4) {\large (b)};
    \end{tikzpicture}
    \caption{Real-world deployment of the UWB TDOA localization system in the multi-room environment (a) based on the optimal sensor placement design. The anchors are enclosed by red circles and the UWB handheld device is highlighted with a blue circle. 
    The estimated results together with the total station ground truth trajectories from all five experiments are summarized in (b). The experimental results show that the propose localization system achieves an average $28$ centimeters positioning accuracy in this multi-room scenario.}
    \label{fig:cafe-hallway-exp}
\end{figure*}
We compute the theoretical lower bound of the localization RMSE along the experimental trajectories through sensor placement analysis and present them together with the ESKF estimation errors over time in the bottom row of Figure~\ref{fig:stair-setup}c. Although the real-world ESKF estimation errors are considerably higher than the theoretical lower bound due to the challenging radio propagation conditions in the staircase, the theoretical values predicted the localization error peaks by analyzing the anchor-tag geometry. The localization performance degrades significantly around $20$ seconds in trajectory $\#1$ and around $30$ and $55$ seconds in trajectory $\#2$, which correspond to the time that we approached the boundary of the anchor constellation. These results demonstrate that sensor placement analysis can identify regions of degraded localization performance, providing valuable information for path planning algorithms of mobile robots, for example.

\subsection{A Case Study: Localization in a Multi-room Scenario}
To demonstrate the entire process of constructing a UWB TDOA localization system, we select a challenging multi-room scenario, which includes a cafeteria and a narrow hallway, as our case study. We selected a cafeteria together with a hallway as the experimental space. The multi-room space dimensions were obtained from the building blueprint with the height measured manually. The 3D layout of the multi-room environment, accurately scaled to reflect the real-world dimensions, is visualized in Figure~\ref{fig:hallway-design}a. We modeled the fridge, counter table, and the vending machine as metal obstacles that will lead to severe NLOS scenarios~\cite{zhao2022finding} and ignored the movable tables, chairs, and sofas in the cafeteria. As an illustrative example, we aimed to design a UWB TDOA localization system with the goal of achieving a theoretical localization RMSE lower bound of $20$~cm along a pre-defined trajectory from the center of the cafeteria to the end of the hallway. We configured the UWB anchors from Bitcraze into decentralized mode~\cite{zhao2024util}, which enables scalability in the number of anchors, to cover the entire space. The anchor positions were constrained to be on the boundary of the 3D space for installation purposes.

We employed the optimal sensor placement algorithm~\cite{zhao2022finding} briefly introduced in Section~\ref{sec:osp} to optimize both the required number of anchors and their corresponding positions to meet the accuracy requirement. A trajectory represented by $22$~sample points are selected and the anchor positions are optimized for them. We assume the UWB LOS measurements are unbiased with the default $10$~cm standard deviation and ignore multi-path effects. The optimization results indicated that $16$~anchors, arranged into $8$~anchor pairs $\Gamma = \{(1,2), (3,4),\cdots, (15, 16)\}$, are required to achieve an average RMSE lower bound of $20$~cm over the sample points. The optimized anchor positions and the corresponding localization accuracy heatmap, with lower RMSE indicated with darker color, are visualized in Figure~\ref{fig:hallway-design}b. With the optimized anchor placement, each sampled point maintains LOS to several anchor pairs in this challenging environment to ensure reliable localization. As Bitcraze's UWB anchors communicate and synchronize among all nearby anchors in decentralized mode, UWB tag will receive out-of-sequence TDOA measurements $d_{ij}\notin \Gamma$. As the out-of-sequence UWB measurements often suffer from degraded anchor geometries, we inflated the variance for $d_{ij}\notin \Gamma$ as $0.025$ in the ESKF algorithm to fuse these measurements. We also increased the Mahalanobis distance parameter of the chi-squared test to $10$ for outlier rejection in this challenging environment. 
\begin{figure}[b!]
    \centering
    \begin{tikzpicture}
    \node[inner sep=0pt, opacity=1.0] (anchor) at (0.0,0.0)
    {\includegraphics[width=.4\textwidth]{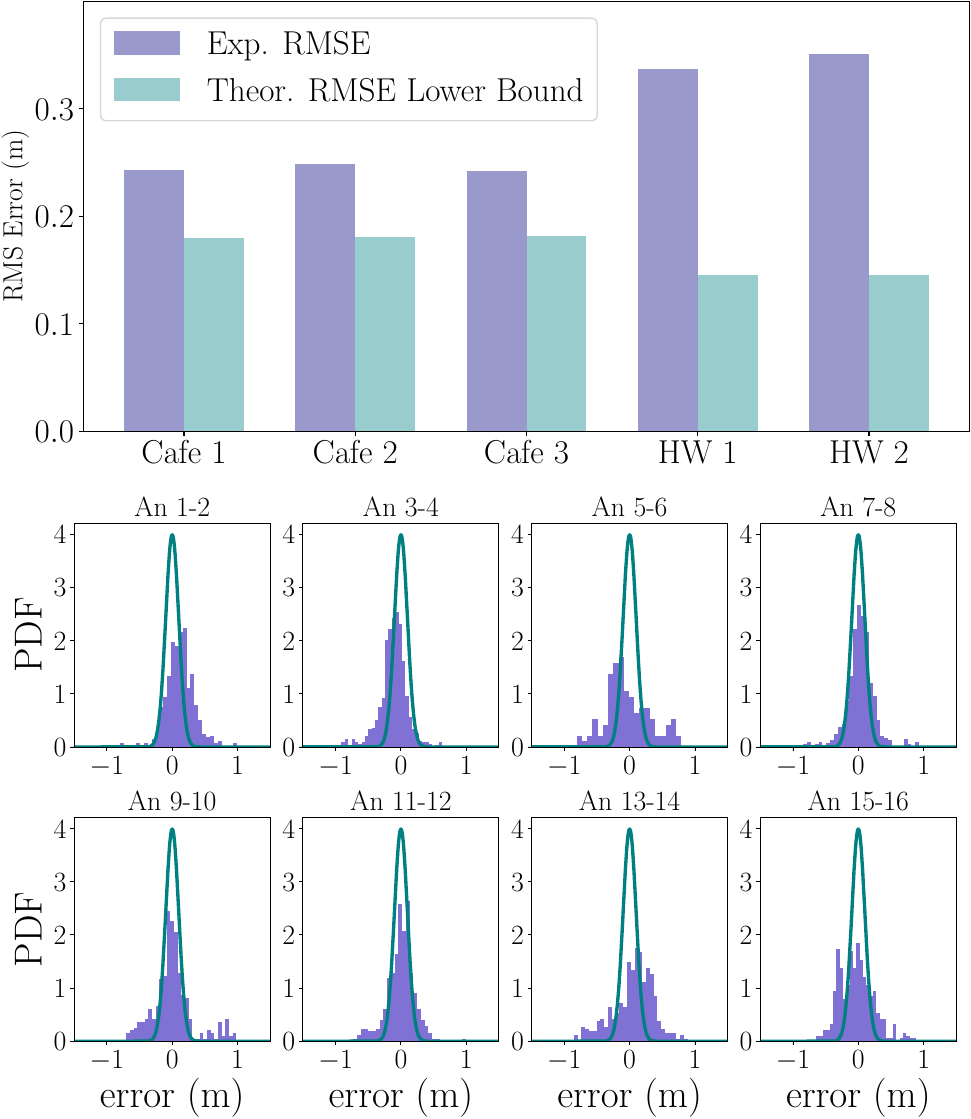}};
    \node[text width=3cm] at (-2.1, 4.0) {\normalsize (a)};
    \node[text width=3cm] at (-2.1, 0.25) {\normalsize (b)};
    \end{tikzpicture}
    \caption{Simulation and experimental root-mean-squared error (RMSE) results (a) and histograms of UWB TDOA measurement residuals in the multi-room environment (b). The default zero-mean Gaussian noise distribution $n_{ij}\sim\mathcal{N}(0,0.01)$ under LOS conditions (green) is overlaid onto the histograms (blue) for comparisons. The UWB measurements are corrupted during the experiments, leading to the gap between the theoretical analysis and experimental results.  }
    \label{fig:cafe-hallway-fig}
\end{figure}

After determining the number of anchors and their positions, we deployed the $16$~UWB anchors in the cafeteria and the hallway according to the sensor placement design. The deployment of the localization system during the experiments is shown in Figure~\ref{fig:cafe-hallway-exp}a. As explained in Section~\ref{sec:arch_deploy}, we constructed the inertial frame and surveyed the anchor positions within this frame using a Leica total station. The average RMSE between the deployed anchor positions compared to the designed positions is $22$~cm, which is partially due to the inconsistency between the actual construction results and the building blueprints. However, this discrepancy is negligible compared to the overall scale of the space, which spans over $25$~m. Due to the challenging geometry at the tight corners around the doorway, the total station lost track at the intersection of the cafeteria and the hallway. Consequently, we had to conduct experiments in the cafeteria and the hallway separately to obtain reliable ground truth position measurements from the Leica total station to quantify the localization accuracy. 

We conducted three experiments in the cafeteria and two experiments in the hallway following a predefined trajectory and summarize all the experimental results  together with the ground truth trajectories in Figure~\ref{fig:cafe-hallway-exp}b. The corresponding RMSE for each experiment and the theoretical RMSE analysis are summarized in Figure~\ref{fig:cafe-hallway-fig}a. The average localization RMSE in the cafeteria and the narrow hallway are $24$~cm and $34$~cm, respectively, leading to an overall average of $28$~cm in this multi-room environment.

Although the localization accuracies, considering the low-cost IMU and UWB sensors we used in this challenging environment, are commendable and sufficient for pedestrian localization applications, there remains a gap between theoretical analysis and experimental results. The average theoretical RMSE lower bound from sensor placement analysis is $17$~cm. This discrepancy primarily arises from mismatches between the assumed UWB noise model and real-world measurements. Figure~\ref{fig:cafe-hallway-fig}b shows histograms of UWB TDOA residuals across five trajectories, overlaid with the assumed zero-mean Gaussian noise model $n_{ij}\sim\mathcal{N}(0,0.01)$ under LOS conditions. In practice, the measurements exhibit significant corruption due to intrinsic DW1000 biases~\cite{zhao2021learning} and multipath propagation, particularly in the narrow hallway—effects that are neglected in the theoretical analysis.

To address this issue, we recommend a conservative sensor placement design by setting a higher standard deviation for LOS measurements within the algorithm. Alternatively, if there exists a method to better simulate and detect multi-path radio propagation in UWB communication, we can incorporate that into the sensor placement optimization to enhance the overall performance of the UWB TDOA localization system.

\section{Conclusion}
\label{sec:conclusion}
In this article, we presented a system-level UWB TDOA localization framework, spanning sensor placement optimization, hardware and software integration, and real-world deployment. The proposed system provides accurate and robust localization while remaining cost-effective, lightweight, and portable. Extensive experiments across diverse environments demonstrate reliable multi-agent pedestrian localization, robustness to complete sensor occlusion, and the ability of sensor placement analysis to predict performance degradation in a geometrically challenging staircase. A multi-room case study involving a cafeteria and a narrow hallway further illustrates the complete deployment pipeline, from placement optimization to experimental validation. In this environment spanning over $25$~m, the system achieves an average positioning accuracy of approximately $28$~cm, highlighting the potential of scalable, low-cost radio-based localization for indoor applications.
\section{Acknowledgement}
\label{sec:acknow}
We would like to thank Adam Heins, Connor Jong, Keenan Burnett, and Sepehr Samavi for their assistance in the experiments. This work was supported in part by the Natural Sciences and Engineering Research Council of Canada (NSERC) and in part by the Canada CIFAR AI Chairs Program.

\bibliographystyle{./IEEEtranBST/IEEEtran}
\bibliography{./IEEEtranBST/IEEEabrv,./biblio}

\end{document}